\documentclass{ieeeaccess}
\usepackage{cite}
\usepackage{amsmath,amssymb,amsfonts}
\usepackage{algorithmic}
\usepackage{graphicx}
\usepackage{textcomp}
\usepackage{adjustbox}
\usepackage{microtype}
\usepackage{times}
\usepackage{latexsym}
\usepackage{booktabs}
\usepackage{multirow}
\usepackage{verbatim}
\usepackage{float}
\usepackage{bbm}
\usepackage{dsfont}
\usepackage{amssymb}
\usepackage{arydshln}
\usepackage{amsmath}
\usepackage{nccmath}
\usepackage{pifont}
\usepackage{newfloat}
\usepackage{enumitem}
\usepackage{hyperref}
\usepackage{subfig}
\usepackage{tabularx}
\usepackage{cleveref}
\usepackage{diagbox}

\def\BibTeX{{\rm B\kern-.05em{\sc i\kern-.025em b}\kern-.08em
    T\kern-.1667em\lower.7ex\hbox{E}\kern-.125emX}}
\begin{document}
\doi{10.1109/ACCESS.2022.3157854}

\newcolumntype{Y}{>{\centering\arraybackslash}X}
\definecolor{goldenpoppy}{rgb}{0.99, 0.76, 0.0}
\definecolor{cadmiumred}{rgb}{0.89, 0.0, 0.13}
\definecolor{ao}{rgb}{0.0, 0.0, 1.0}
\definecolor{ao(english)}{rgb}{0.0, 0.5, 0.0}

\newcommand{\etal}{\textit{et al.}}

\newcommand{\dicttrain}{DICT$_\text{train}$}
\newcommand{\dictsyn}{DICT$_\text{syn}$}
\newcommand{\bert}{BERT}
\newcommand{\biobert}{BioBERT}
\newcommand{\bluebert}{BlueBERT}
\newcommand{\pubmedbert}{PubMedBERT}

\newcommand{\ncbi}{NCBI-disease}
\newcommand{\cdr}{BC5CDR}
\newcommand{\cdrdis}{BC5CDR$_\text{dis}$}
\newcommand{\cdrchem}{BC5CDR$_\text{chem}$}

\newcommand{\mem}{\texttt{Mem}}
\newcommand{\syn}{\texttt{Syn}}
\newcommand{\con}{\texttt{Con}}
\newcommand{\tmem}{\texttt{\textbf{Mem}}}
\newcommand{\tsyn}{\texttt{\textbf{Syn}}}
\newcommand{\tcon}{\texttt{\textbf{Con}}}

\newcommand{\precision}{P}
\newcommand{\recall}{R}
\newcommand{\fscore}{F$_\text{1}$}

\newcommand{\smallspace}{$~\;$}
\newcommand{\tinyspace}{$~$}


\title{How Do Your Biomedical Named Entity Recognition Models Generalize to \\Novel Entities?}
\author{\uppercase{Hyunjae Kim}\authorrefmark{1} and \uppercase{Jaewoo Kang}\authorrefmark{1}}
\address[1]{Department of Computer Science and Engineering, Korea University, Seoul, South Korea}
\tfootnote{
This research was supported by National Research Foundation of Korea (NRF-2020R1A2C3010638), a grant of the Korea Health Technology R\&D Project through the Korea Health Industry Development Institute (KHIDI), funded by the Ministry of Health \& Welfare, Republic of Korea (grant number: HR20C0021), and the MSIT (Ministry of Science and ICT), Korea, under the ICT Creative Consilience program (IITP-2022-2020-0-01819) supervised by the IITP (Institute for Information \& communications Technology Planning \& Evaluation).}

\markboth
{Kim and Kang: How Do Your Biomedical Named Entity Recognition Models Generalize to Novel Entities?}
{Kim and Kang: How Do Your Biomedical Named Entity Recognition Models Generalize to Novel Entities?}

\corresp{Corresponding author: Jaewoo Kang (e-mail: kangj@korea.ac.kr).}

\begin{abstract}
The number of biomedical literature on new biomedical concepts is rapidly increasing, 
which necessitates a reliable biomedical named entity recognition (BioNER) model for identifying new and unseen entity mentions.
However, it is questionable whether existing models can effectively handle them.
In this work, we systematically analyze the three types of recognition abilities of BioNER models: memorization, synonym generalization, and concept generalization.
We find that although current best models achieve state-of-the-art performance on benchmarks based on overall performance, they have limitations in identifying synonyms and new biomedical concepts, indicating they are overestimated in terms of their generalization abilities.
We also investigate failure cases of models and identify several difficulties in recognizing unseen mentions in biomedical literature as follows: 
(1) models tend to exploit dataset biases, which hinders the models’ abilities to generalize, and (2) several biomedical names have novel morphological patterns with weak name regularity, and models fail to recognize them.
We apply a statistics-based debiasing method to our problem as a simple remedy and show the improvement in generalization to unseen mentions.
We hope that our analyses and findings would be able to facilitate further research into the generalization capabilities of NER models in a domain where their reliability is of utmost importance.\footnote{Code and datasets are available at \href{https://github.com/dmis-lab/bioner-generalization}{https://github.com/dmis-lab/bioner-generalization}.}

\end{abstract}

\maketitle

\section{Introduction}

Recently, more than 3,000 biomedical papers are being published per day on average \cite{landhuis2016scientific, wang2020cord}.
Searching these documents efficiently or extracting useful information from them would be of great help to researchers and practitioners in the field.
Biomedical named entity recognition (BioNER), which involves identifying biomedical named entities in unstructured text, is a core task to do so since entities extracted by BioNER systems are utilized as important features in many downstream tasks such as drug-drug interaction extraction \cite{liu2016drug}.

One important desideratum of BioNER models is to be able to generalize to unseen entity mentions.
This generalization capability is of paramount importance in the biomedical domain due to the following reasons.
First, various expressions for a biomedical entity (i.e., synonyms) continue to be made.
For instance, pharmaceutical companies come up with marketing-appropriate names such as Gleevec to replace old names (usually identifiers) such as CGP-57148B and STI-571, whereas entities in other domains such as countries and companies are relatively unchanged.
Second, new biomedical entities and concepts such as the novel coronavirus disease 2019 (COVID-19) constantly emerge, which can have a direct impact on human life and health.

In contrast to the importance of generalizing to new entities in the biomedical literature, there has been little systematic analysis of the generalizability of BioNER models.
While recent works have made great efforts to push the state-of-the-art (SOTA) on various benchmarks \cite{habibi2017deep, wang2019cross, yoon2019collabonet, lee2020biobert}, it is questionable whether a high overall performance on a benchmark indicates true generalization.
We conducted a pilot study to check if current BioNER models are reliable in identifying new entities.
Specifically, we trained \biobert~\cite{lee2020biobert} on the NCBI corpus \cite{dougan2014ncbi}, and then tested how many spans containing the novel entity COVID-19 the model can extract from PubMed sentences. 
As a result, the model extracted only 45.7\% of all the spans, although it achieved high overall performance on NCBI (90.5\% in recall).
From this, we conclude that existing BioNER models may have limitations in identifying unseen entities, and their generalizability should be explored in a more systematic way beyond measuring overall performance.

\begin{table*}[t]
\centering
\normalsize
\caption{
The number of mentions in the \mem, \syn, and \con~splits of benchmarks.
Each split (i.e., \mem, \syn, and \con) corresponds to each recognition type (i.e., memorization, synonym generalization, and concept generalization).
The table shows that current BioNER benchmarks are overrepresented by the mentions in the \mem~splits (i.e., memorizable mentions).
}
\label{tab:statistics}
\begin{adjustbox}{max width = 0.95\textwidth}
\begin{tabular}{llcccccc}
\toprule
  &  &\multicolumn{3}{c}{\textbf{Validation}} & \multicolumn{3}{c}{\textbf{Test}} \\
 \cmidrule(lr){3-5} \cmidrule(l){6-8}
\textbf{Dataset} & \textbf{Type} & \tmem & \tsyn & \tcon & \tmem & \tsyn & \tcon \\
\midrule
\ncbi & Disease &   \begin{tabular}[c]{@{}r@{}} \multicolumn{1}{c}{599} \\ (62.4\%)  \end{tabular} & \begin{tabular}[c]{@{}c@{}} \multicolumn{1}{c}{196} \\ (20.4\%) \end{tabular} & \begin{tabular}[c]{@{}c@{}} \multicolumn{1}{c}{165} \\ (17.2\%)  \end{tabular} &   \begin{tabular}[c]{@{}c@{}} \multicolumn{1}{c}{515} \\ (65.4\%)  \end{tabular} & \begin{tabular}[c]{@{}c@{}} \multicolumn{1}{c}{191} \\ (24.3\%) \end{tabular} & \begin{tabular}[c]{@{}c@{}} \multicolumn{1}{c}{81} \\ (10.3\%)  \end{tabular} \\
\cdrdis & Disease  & \begin{tabular}[c]{@{}c@{}} \multicolumn{1}{c}{2,807} \\ (63.4\%) \end{tabular} & \begin{tabular}[c]{@{}c@{}} \multicolumn{1}{c}{922} \\ (20.8\%) \end{tabular} & \begin{tabular}[c]{@{}c@{}} \multicolumn{1}{c}{695} \\ (15.7\%) \end{tabular}  &   \begin{tabular}[c]{@{}c@{}} \multicolumn{1}{c}{2,642} \\ (62.3\%)  \end{tabular} & \begin{tabular}[c]{@{}c@{}} \multicolumn{1}{c}{960} \\ (22.6\%) \end{tabular} & \begin{tabular}[c]{@{}c@{}} \multicolumn{1}{c}{642} \\ (15.1\%)  \end{tabular} \\
\cdrchem & Drug/chem. &   \begin{tabular}[c]{@{}c@{}} \multicolumn{1}{c}{3,294} \\ (61.2\%) \end{tabular} & \begin{tabular}[c]{@{}c@{}} \multicolumn{1}{c}{510} \\ (9.5\%) \end{tabular} & \begin{tabular}[c]{@{}c@{}} \multicolumn{1}{c}{1,581} \\ (29.4\%) \end{tabular} &   \begin{tabular}[c]{@{}c@{}} \multicolumn{1}{c}{3,438} \\ (64.3\%)  \end{tabular} & \begin{tabular}[c]{@{}c@{}} \multicolumn{1}{c}{456} \\ (8.5\%) \end{tabular} & \begin{tabular}[c]{@{}c@{}} \multicolumn{1}{c}{1,453} \\ (27.2\%)  \end{tabular} \\
\bottomrule
\end{tabular}
\end{adjustbox}
\end{table*}

In this work, we analyze how well existing BioNER models generalize to unseen mentions.
First, we define three types of recognition abilities that BioNER models should possess:
\begin{itemize}[noitemsep]
    \item \textbf{Memorization}: 
    The most basic ability is to identify the entity mentions that were seen during training.
    We call this type of mention \textit{memorizable} mention.
    If there is no label inconsistency, even a simple rule-based model would recognize memorizable mentions easily.
    \item \textbf{Synonym generalization}: 
    Biomedical names are expressed in various forms, even when they refer to the same biomedical concepts. 
    For instance, Motrin and Ibuprofen are the same entity, but their surface forms are highly different \cite{sung-etal-2020-biomedical}.
    A BioNER model should be robust to these morphological variations (i.e., synonyms).
    \item \textbf{Concept generalization}: 
    While synonym generalization deals with recognizing new surface forms of existing entities, concept generalization refers to the generalization to novel entities or concepts that did not exist before.
    New biomedical concepts such as COVID-19 sometimes are very different from conventional entities in terms of their surface forms and the context in which they appear, which makes it difficult to identify them.
\end{itemize}

In terms of the three recognition types that we define, we partition the entity mentions in the test sets (or validation sets) into three splits based on mention and CUI (Concept Unique Identifier) overlaps with the training sets, as shown in \Cref{tab:statistics}.
This gives us several advantages.
First, we can compare models' generalization abilities in detail.
For instance, we find that the gap in performance between \biobert~and \bert~\cite{devlin-etal-2019-bert} is mainly from synonym and concept generalization, not memorization (\Cref{sec:measure_the_generalization}).
Also, our classification is simple and can be easily adopted to other datasets and other downstream tasks in the biomedical domain such as relation extraction and normalization.
We focus on two popular BioNER benchmark in this work: \ncbi~\cite{dougan2014ncbi} and \cdr~\cite{li2016biocreative}.

On the three test splits, we investigate the generalizability of existing BioNER models.
Despite their SOTA performance on the benchmarks, they have limitations in their generalizability.
Specifically, the models perform well on memorizable mentions, but find it difficult to identify unseen mentions.
For the disease mentions in the \cdr~corpus, \biobert~achieved a recall of 93.3\% on memorizable mentions, but 74.9\% and 73.7\% on synonyms and new concepts, respectively.
Also, the models cannot recognize the newly emerging biomedical concept COVID-19 well. 
Surprisingly, \biobert~recognized only 3.4\% spans containing COVID-19 when trained on \cdr.
From these observations, we conclude that existing BioNER models achieved high performance on benchmarks, but they are overestimated in terms of their generalizability.

Also, we identify several difficulties in recognizing unseen mentions.
First, through a qualitative analysis of error cases on \syn~and \con~splits, we find BioNER models can rely on the class distributions of each word in the training set, reducing the models' abilities to generalize.
Since BioNER datasets is relatively small for training large neural networks, models may be sensitive to such dataset bias.
Second, after examining the failure for COVID-19, we conclude models are not robust to new entities when they do not follow conventional surface patterns.
This is an important issue to be addressed since many biomedical entities have rare morphologies (See \Cref{table:name_reg_results} for examples), and such entities will continue to appear in biomedical literature.

The two difficulties can be viewed as models' biases on statistical cues and surface patterns.
In order to show they are addressable, we apply a simple statistics-based debiasing method~\cite{ko-etal-2020-look}.
Specifically, we use the class distributions of words in the training set as bias prior distributions.
This reduces the training signals from words whose surface forms are very likely to be entities (or non-entities), mitigating models' bias on class distributions and name regularity.
In experiments, we demonstrate our debiasing method consistently improves the generalization to synonyms, new concepts, and entities with unique forms including COVID-19.

To sum up, we make the following contributions:
\begin{itemize}[noitemsep, topsep=0pt]
    \item We first define memorization, synonym generalization, and concept generalization and systematically investigate existing BioNER models in this regard.
    \item We raise the overestimation issue in terms of BioNER models' generalizability to unseen mentions and provide empirical evidence to support our claim.
    \item We identify two types of bias as the main difficulty in generalization in BioNER and show that they are addressable using a current debiasing method.
\end{itemize}

\section{Data Preparation}

\subsection{Partitioning Benchmarks}
\label{subsec:partitioning}

We describe how we partition benchmarks.
Several BioNER datasets provide entity mentions and also CUIs that link the entity mentions to their corresponding biomedical concepts in databases.
We utilize overlaps in mentions and CUIs between training and test sets in the partitioning process.
Let $(\mathrm{x}_n, \mathrm{e}_n, \mathrm{c}_n)$ be the $n$-th data example of a total of $\mathit{N}$ examples in a test set.
$\mathrm{x}_n$ is the $n$-th sentence, 
$\mathrm{e}_n = [e_{(n,1)}, ..., e_{(n,\mathit{T}_n)}]$ is a list of entity mentions, 
and $\mathrm{c}_n$ = $[c_{(n,1)}$, ..., $c_{(n,\mathit{T}_n)}]$ is a list of CUIs where $\mathit{T}_n$ is the number of the entity mentions (or CUIs) in the sentence.
We partition all mentions $e_{(n,t)}$ in the original test set into three splits as follows:
\begin{alignat*}{2}
    \boldsymbol{\cdot}\ & \mathrm{\mem} && := \left\{ e_{(n, t)}: e_{(n, t)} \in \mathbb{E}_\text{train}, c_{(n, t)} \in \mathbb{C}_{\text{train}} \right\} \\
    \boldsymbol{\cdot}\ & \mathrm{\syn} && := \left\{ e_{(n, t)}: e_{(n, t)} \notin \mathbb{E}_\text{train}, c_{(n, t)} \in \mathbb{C}_{\text{train}} \right\} \\
    \boldsymbol{\cdot}\ & \mathrm{\con} && := \left\{ e_{(n, t)}: e_{(n, t)} \notin \mathbb{E}_\text{train}, c_{(n, t)} \notin \mathbb{C}_{\text{train}} \right\}
\end{alignat*}
where $\mathbb{E}_\text{train}$ is the set of all entity mentions in the training set, and $\mathbb{C}_\text{train}$ is the set of all CUIs in the training set.
We describe the partitioning process in detail in the Appendix.

\subsection{Datasets}

We use two popular BioNER benchmarks with CUIs to systematically investigate models' memorization, synonym generalization, and concept generalization abilities.
Additionally, we automatically construct a dataset consisting of the novel entity COVID-19.

\subsubsection{\ncbi}
The \ncbi~corpus \cite{dougan2014ncbi} is a collection of 793 PubMed articles with manually annotated disease mentions and the corresponding concepts in Medical Subject Headings (MeSH) or Online Mendelian Inheritance in Man (OMIM).

\subsubsection{\cdr}
The \cdr~corpus \cite{li2016biocreative} is proposed for disease name recognition and chemical-induced disease (CID) relation extraction tasks.
The corpus consists of 1,500 manually annotated disease and chemical mentions and the corresponding concepts in MeSH.
We denote the disease-type dataset as \cdrdis~and the chemical-type dataset as \cdrchem.

\subsubsection{COVID-19}
We construct a dataset to see if a model trained on current benchmarks can identify the newly emerging biomedical concept COVID-19.
We sampled 5,000 sentences containing ``COVID-19'' from the entire PubMed abstracts through March 2021 and annotated all COVID-19 occurrences in the sentences, which results in 5,237 labels. 
Note that only the exact term “COVID-19” was considered, and synonyms for COVID-19 were not considered in this dataset creation process.

\subsection{Split Statistics}
\label{subsec:statistics}
\Cref{tab:statistics} shows the statistics of the splits of the benchmarks.
We found that a significant portion of the benchmarks correspond to \mem, implying that current BioNER benchmarks are highly skewed to memorizable mentions.
In \Cref{sec:measure_the_generalization}, we discuss the overestimation problem that such overrepresentation of memorizable mentions may cause.

\section{Generalizability of BioNER Models}
\label{sec:measure_the_generalization}

\begin{table*}[t]
\centering
\normalsize
\caption{
Performance of current BioNER models on \ncbi, \cdrdis, and \cdrchem. 
The best scores are highlighted in bold, and the second best scores are underlined.
}
\label{tab:main_result}
\begin{tabular}{ll|rrr|rrr|l}
\toprule
\multirow{2}{*}{\diagbox[width=9em]{\textbf{Training}}{\textbf{Test}}} & \multicolumn{1}{c}{} & \multicolumn{3}{c}{\textbf{Overall}} & \multicolumn{3}{c}{\textbf{In-depth}} & \multirow{3}{*}{\begin{tabular}[c]{@{}c@{}} {\textbf{COVID}} \\ {\textbf{-19}} \end{tabular}} \\
\cmidrule(lr){3-5} \cmidrule(lr){6-8}
 & \multicolumn{1}{l}{\textbf{Model}} & \multicolumn{1}{c}{\textbf{\precision}} & \multicolumn{1}{c}{\textbf{\recall}} & \multicolumn{1}{c}{\textbf{\fscore}} & \multicolumn{1}{c}{\textbf{\mem}} & \multicolumn{1}{c}{\textbf{\syn}} & \multicolumn{1}{c}{\textbf{\con}} & \\
\midrule
\multirow{6}{*}{\ncbi} & \pubmedbert~\cite{gu2020domain} & \underline{86.6} & 88.9 & \underline{87.7} & 94.5 & \textbf{81.1} & \underline{77.7} & \smallspace\underline{36.0}  \\
 & \bluebert~\cite{peng2019transfer} & 85.8 & \underline{89.5} & 87.6 & \textbf{95.9} & 80.4 & 76.7 & \smallspace13.8 \\
 & \biobert~\cite{lee2020biobert} & \textbf{86.7} & \textbf{90.5} & \textbf{88.6} & 95.5 & \underline{80.9} & \textbf{84.1} & \smallspace\textbf{45.7} \\
 & \bert~\cite{devlin-etal-2019-bert} & 83.8 & 87.6 & 85.6 & 95.0 & 79.0 & 70.7 & \smallspace18.7 \\
 \cmidrule(l){2-9}
 & \dictsyn & 50.7 & 58.0 & 54.1 & 88.5 & 13.8 & 0.0 & \smallspace\smallspace0.0 \\
 & \dicttrain & 52.7 & 55.4 & 54.0 & 88.8 & 0.0 & 0.0 & \smallspace\smallspace0.0 \\
\midrule
\multirow{6}{*}{\cdrdis} & \pubmedbert~\cite{gu2020domain} & \textbf{83.1} & \textbf{87.3} & \textbf{85.2} & 93.1 & \textbf{78.3} & \textbf{75.7} & \smallspace\smallspace\underline{2.2}  \\
 & \bluebert~\cite{peng2019transfer} & 82.2 & \underline{86.6} & \underline{84.4} & 93.2 & \underline{76.9} & 72.7 & \smallspace\smallspace0.8 \\
 & \biobert~\cite{lee2020biobert} & \underline{82.4} & 86.3 & 84.3 & \underline{93.3} & 74.9 & \underline{73.7} & \smallspace\smallspace\textbf{3.4} \\
 & \bert~\cite{devlin-etal-2019-bert} & 78.5 & 81.4 & 79.9 & 91.5 & 64.0 & 63.4 & \smallspace\smallspace0.8 \\
 \cmidrule(l){2-9}
 & \dictsyn & 75.4 & 67.8 & 71.4 & 96.0 & 32.9 & 0.0 & \smallspace\smallspace0.0  \\
 & \dicttrain & 75.9 & 61.4 & 67.8 & 96.7 & 0.0 & 0.0 & \smallspace\smallspace0.0  \\
\midrule
\multirow{6}{*}{\cdrchem} & \pubmedbert~\cite{gu2020domain} & 92.1 & \textbf{94.2} & \textbf{93.1} & \textbf{98.3} & \textbf{85.5} & \textbf{88.2} & \multicolumn{1}{c}{-} \\
 & \bluebert~\cite{peng2019transfer} & \textbf{92.8} & 92.9 & \underline{92.8} & 98.0 & 81.6 & 86.0 & \multicolumn{1}{c}{-}  \\
 & \biobert~\cite{lee2020biobert} & 92.1 & \underline{93.1} & 92.6 & 97.8 & \underline{82.1} & \underline{87.0} & \multicolumn{1}{c}{-} \\
 & \bert~\cite{devlin-etal-2019-bert} & 89.8 & 90.0 & 89.9 & 97.0 & 72.4 & 81.1 & \multicolumn{1}{c}{-} \\
 \cmidrule(l){2-9}
 & \dictsyn & 71.5 & 62.2 & 66.5 & 95.9 & 32.7 & 1.4 & \multicolumn{1}{c}{-}  \\
 & \dicttrain & 71.2 & 58.8 & 64.6 & 96.2 & 0.0 & 0.0 & \multicolumn{1}{c}{-} \\
\bottomrule
\end{tabular}
\end{table*}

This section describes baseline models and evaluation metrics and analyzes the three recognition abilities of the models.

\subsection{Baseline Models}
We use four current best neural net-based models and two traditional dictionary-based models as our baseline models.
See the Appendix for implementation details.

\subsubsection{Neural Models}
We use \biobert~\cite{lee2020biobert}, \bluebert~\cite{peng2019transfer}, and \pubmedbert~\cite{gu2020domain}.
The models are all pretrained language models (PLMs) for the biomedical domain, with similar architectures.
They are different in their vocabularies, weight initialization, and training corpora.
See the Appendix for more details.
Also, we use \bert~\cite{devlin-etal-2019-bert} to compare general and domain-specific PLMs in terms of generalization in BioNER.

\subsubsection{Dictionary Models}
Traditional approaches in the field of BioNER are based on pre-defined dictionaries \cite{rindflesch1999edgar}.
To compare the generalization abilities of traditional and recent approaches, we set two types of simple dictionary-based extractors as baseline models.
\dicttrain~uses all the entity mentions in a training set (i.e., $\mathbb{E}_\text{train}$) as a dictionary and classifies text spans as entities when the dictionary includes the spans.
If candidate spans overlap, the longest one is selected.
\dictsyn~expands the dictionary to use entity mentions in the training set as well as their synonyms, which are pre-defined in biomedical databases.

\subsection{Metrics}
Following conventional evaluation metrics in BioNER, we use the precision (\precision), recall (\recall), and F1 score (\fscore) at an entity level to measure overall performance~\cite{tjong-kim-sang-de-meulder-2003-introduction}.
We only use recall when evaluating three recognition abilities (i.e., \mem,~\syn, and~\con) since it is impossible to classify false positives into each recognition type.
For COVID-19, we use a relaxed version of recall: if ``COVID-19'' is contained in the predicted spans, we classify this prediction as a true positive.

\subsection{Results}

\subsubsection{Overall Results}
\Cref{tab:main_result} shows the performance of the baseline models.
\biobert~outperforms other baseline models on \ncbi~based on overall performance.
For the \cdr~corpus, \pubmedbert~is the best performing model.
BERT performs less than domain-specific PLMs, but far superior to dictionary models.
\dictsyn~outperforms \dicttrain~in recall due to its larger biomedical dictionary, but the precision scores decrease in general.
Note that the performance of \dictsyn~on \mem~is lower than that of \dicttrain~as there exists annotation inconsistency between benchmarks and biomedical databases.
We elaborate on this in the Appendix.

Memorization can be easily obtained compared to the other two abilities.
Although the dictionary models are the simplest types of BioNER models without learnable parameters, they work well on \mem.
The degree of difficulty in recognizing synonyms and new concepts varies from data to data.
The models' performances on \syn~is lower than that on \con~of \cdrdis, but vice-versa on \cdrchem.


\subsubsection{Overestimation of Models}
\label{para:overestimation}

The neural models perform well on \mem, but they achieved relatively low performance on \syn~and \con~across all benchmarks.
For instance, \biobert~achieved 93.3\% recall on \mem, but only 74.9\% and 73.7\% recall on \syn~and \con, respectively.
Also, the neural models perform poorly on COVID-19 despite their high F1 scores.
\biobert~performed the best, but the score is only 45.7\% recall.
Even more surprisingly, all the models hardly identify COVID-19 when trained on \cdrdis.
To sum up, current BioNER models have limitations in their generalizability.

As shown in \Cref{tab:statistics}, a large number of entity mentions in existing BioNER benchmarks are included in \mem.
This overrepresentation of memorizable mentions can lead to an overestimation of the generalization abilities of models. 
We believe our model has high generalization ability due to high performance on benchmarks, but the model may be highly fit to memorizable mentions.
Taking these results into account, we would like to emphasize that researchers should be wary of falling into the trap of overall performance and misinterpreting a model's high performance with generalization ability at the validation and inference time.

\subsubsection{Effect of Domain-specific Pretraining}
\label{subsec:main_result}

Domain-specific PLMs constantly outperform \bert~on \syn~and \con.
These results show that pretraining on domain-specific corpora mostly affects synonym generalization and concept generalization.
On the other hand, \bert~and domain-specific PLMs achieve similar performance on \mem~because memorization does not require much domain-specific knowledge and the models have the same architecture and capacity.

\begin{table}[t]
\caption{Performance of neural models on the abbreviations in the \con~splits.
32.7\% of mentions are abbreviations in \con~of \ncbi, while \cdrdis~has only 7.2\% abbreviations.
The best scores are highlighted in bold.
}
\label{tab:abbreviation_results}
\centering
\normalsize
\begin{tabular}{lcc}
\toprule
\begin{tabular}[c]{@{}c@{}}\\ \textbf{Model} \end{tabular}   & \begin{tabular}[c]{@{}c@{}}\textbf{\ncbi}\\ (32.7\%) \end{tabular}  & \begin{tabular}[c]{@{}c@{}}\textbf{\cdrdis}\\ (7.2\%) \end{tabular} \\
\midrule
\pubmedbert~\cite{gu2020domain} & 84.8  & \textbf{71.5}  \\
\bluebert~\cite{peng2019transfer} & 81.5  & 70.4  \\
\biobert~\cite{lee2020biobert}  & \textbf{89.6} & 69.6 \\
\midrule
\bert~\cite{devlin-etal-2019-bert} & 65.2 & 52.8        \\ 
\bottomrule
\end{tabular}
\end{table}

In particular, we find the gap in performance between \bert~and domain-specific PLMs is drastic in the generalization ability to abbreviations.
\Cref{tab:abbreviation_results} shows that neural models' performances on abbreviations on the \con~splits of \ncbi~and \cdrdis.\footnote{Note that~\cdrchem~is excluded in this experiment since it is not easy to distinguish between abbreviations and other chemical entities such as identifiers and formula due to their similar forms.}
On \ncbi, \biobert~is very robust to abbreviations, and the gap in performance between \biobert~and \bert~is 24.4\% in recall.
\biobert~also significantly outperforms the other domain-specific PLMs, resulting in high performance on \con~of \ncbi.
On the other hand, \pubmedbert~is the best on the \cdrdis, outperforming \bert~by a recall of 19.7\%.

\section{Analysis}
\label{sec:method}
In this section, we analyze which factors make the generalization to unseen biomedical names difficult based on failures of models on (1) \syn~and \con~splits, and (2) COVID-19.
For simplicity, we will focus on only \bert~and \biobert.

\subsection{Dataset Bias}
\label{subsec:evidence_of_bias}

We qualitatively analyze the error cases of \biobert~by sampling a total of 100 incorrect predictions from the \syn~and \con~splits of \cdrdis.
As a result, we found 36\% of the error cases occur because the model tends to rely on statistical cues in the dataset and make biased predictions.
\Cref{table:evidence_of_bias} shows the examples of the biased predictions.

\begin{table}[t]
\caption{
Examples of biased predictions of \biobert. 
Entity mentions (ground-truth labels) are displayed in blue. 
Model predictions are highlighted with yellow boxes.
}
\label{table:evidence_of_bias}
\centering
\normalsize
\begin{adjustbox}{max width = \columnwidth}

\begin{tabular}{l}
\toprule
\textbf{Example} \\ 
\midrule
\begin{tabular}[c]{@{}l@{}} 
\relax[1] \tinyspace\textbf{{\textcolor{ao}{Acute \colorbox{yellow}{encephalopathy}}}} and \textbf{{\textcolor{ao}{\colorbox{yellow}{cerebral vasospasm}}}} after \\
$\quad$\smallspace multiagent chemotherapy $\dots$
\\
\relax[2] \tinyspace$\dots$ 14 with \colorbox{yellow}{\textbf{{\textcolor{ao}{anterior infarction}}}} (\textbf{{\textcolor{ao}{ANT-\colorbox{yellow}{MI}}}}) and eight \\
$\quad$\smallspace with \colorbox{yellow}{\textbf{{\textcolor{ao}{inferior infarction}}}} (\textbf{{\textcolor{ao}{INF-\colorbox{yellow}{MI}}}}). \\
\relax[3] \tinyspace Two patients needed a lateral \colorbox{yellow}{\textbf{tarsorrhaphy}} for persistent \\ 
$\quad$\smallspace \textbf{{\textcolor{ao}{epithelial defects}}}. 
\end{tabular} \\ 

\bottomrule
\end{tabular}
\end{adjustbox}
\end{table}

In the first example, the model failed to extract the whole phrase ``acute encephalopathy.''
All the words ``encephalopathy'' in the training set are labeled as ``B,''\footnote{\textit{Beginning} in the BIO tagging scheme \cite{ramshaw1999text,tjong-kim-sang-de-meulder-2003-introduction}.} so the model classified the word as ``B,'' resulting in an incorrect prediction.
In the second example, there are four entity mentions: two mentions are full names ``anterior infarction'' and ``inferior infarction,'' and the others are their corresponding abbreviations ``ANT-MI'' and ``INF-MI.''
As the abbreviations are enclosed in parentheses after the full names, it should be easy for a model to identify the abbreviations in general if the model can extract the full names.
Interestingly, although \biobert~correctly predicted the full names in the example, it failed to recognize their abbreviations.
This is because ``MI'' is only labeled as ``B'' in the training set, and so the model was convinced that ``MI'' is only associated with the label ``B.''
In the last example, about 73\% of the words ``defects'' are labeled as ``I'' in the training set as components of entity mentions such as birth defects and atrial septal defects.
However, the word ``epithelial'' is only labeled as ``O,'' so the model did not predict the phrase ``epithelial defects'' as an entity.

From these observations, we hypothesize that BioNER models are biased to class distributions in datasets.
Specifically, models tend to over-rely on the class distributions of each word in the training set, causing the models to fail when the class distribution shifts in the test set.

\subsection{Weak Name Regularity}
\label{subsec:name_regularity}
Name regularity refers to patterns in the surface forms of entities \cite{lin-etal-2020-rigorous,ghaddar2021context}.
For example, many disease names have patterns such as ``$\rule{0.3cm}{0.15mm}$ disease'' and ``$\rule{0.3cm}{0.15mm}$ syndrome.''
These patterns are regarded as useful features for identifying unseen mentions and are often implemented in NER systems after being handcrafted.
However, little analysis has been done on the difficulties a model can face when extracting novel entities that do not have common name patterns such as COVID-19.
In this section, we hypothesize that the cause of models' failure to recognize COVID-19 is its rare morphology and perform detailed analyses to support the hypothesis.

\subsubsection{Cause of Failing to Recognize COVID-19}

\begin{table}[t]
\caption{Performance of \biobert~on COVID-19 and synthetically generated mention ``COVID.''}
\label{tab:edit_covid_form}
\centering
\normalsize
\begin{tabular}{lcl}
\toprule
\diagbox[width=7.5em]{\textbf{Training}}{\textbf{Test}}  & \textbf{COVID-19} & \multicolumn{1}{c}{\textbf{COVID}} \\
\midrule
\ncbi & 45.7 & 85.8 \textcolor{ao}{(+ 40.1)}  \\
\cdrdis  & \smallspace3.4 & 55.1 \textcolor{ao}{(+ 51.7)}  \\
\bottomrule
\end{tabular}
\end{table}

\begin{table}[t]
\caption{Performance of \biobert~on the {COVID}-19 dataset when trained with name patterns similar to COVID-19.}
\label{tab:augment_results}
\centering
\normalsize
\begin{tabular}{lll}
\toprule
\diagbox[width=8em]{\textbf{Model}}{\textbf{Training}} & \multicolumn{1}{l}{\textbf{\ncbi}} & \multicolumn{1}{l}{\textbf{\cdrdis}}  \\
\midrule
BioBERT & 45.7 & \smallspace3.4      \\
\quad+ Replace. (1) & 51.0 \textcolor{ao}{(+ 5.3)} & 14.8 \textcolor{ao}{(+ 11.4)} \\
\quad+ Replace. (10) & 56.6 \textcolor{ao}{(+ 10.9)} & 27.6 \textcolor{ao}{(+ 24.2)}  \\
\bottomrule
\end{tabular}
\end{table}

We have already confirmed in \Cref{tab:main_result} that models fail to recognize COVID-19.
To see if the cause of this failure is due to the rare surface form of COVID-19, we replace all occurrences of ``COVID-19'' in the COVID-19 dataset with more \textit{disease-like} mentions ``COVID,'' while maintaining context.
Interestingly, \biobert~can recognize the entity well after the replacement, as shown in \Cref{tab:edit_covid_form}.

Next, we train models with entity mentions having similar surface forms to COVID-19 and see how the performance changes on COVID-19.
First, we randomly generate 3-5 capital letters and 1-3 numbers. We then combine the generated letters and numbers using the pattern “{Abbreviation}-{Number}” and create pseudo entities such as IST-5, CHF-113, and SRS-3517. 
We randomly select 1 or 10 entity mentions in the training set that are abbreviations and replace them with different pseudo entities.
We then trained \biobert~on the modified training set and test the model how well it recognizes COVID-19.
As shown in \Cref{tab:augment_results}, augmenting COVID-19-like name patterns improves the ability to recognize COVID-19.

Note that low performance on COVID-19 is not due to lack of sufficient context.
Models fail even if there is enough information in the context to determine whether COVID-19 is a disease, e.g., ``\textit{treatment of COVID-19 patients with hypoxia}'' and ``\textit{The 2019 novel coronavirus pneumonia (COVID-19) is an ongoing global pandemic with a worldwide death toll}.''
Also, the small number of training data is not the cause for the failure.
We trained \biobert~on the MedMentions corpus~\cite{mohan2019medmentions}, which contains several times more disease mentions than \ncbi~and \cdrdis, but the model extracted only 12.7\% of COVID-19.
From these observations, we conclude that the biggest difficulty in recognizing COVID-19 is the generalization to a novel surface form.

\subsubsection{Comparison of \ncbi~and \cdr}

When trained on \ncbi~and \cdrdis, the gap in performance between the models on COVID-19 is remarkable.
This can be caused by three factors.
First, the \cdr~corpus contains a number of chemical mentions with the pattern ``\{\textit{Abbreviation}\}-\{\textit{Number}\}'' such as ``MK-486'' and ``FLA-63,'' thus models can misunderstand the pattern must be the chemical type, not a disease type.
Second, \ncbi~contains several times more abbreviations than \cdrdis~in the training set, which could help generalization to COVID-19 that is also an abbreviation.
Lastly, \ncbi~has the entity ``EA-2'' in the training set with a similar pattern to COVID-19, while \cdrdis~does not have any disease entity with the pattern.
Replacing ``EA-2'' with ``EA'' significantly reduces the performance of BioBERT dramatically decreases from 45.7 to 11.2, which supports our claim.

\subsection{Debiasing Method}

We hypothesize BioNER models tend to rely on class distributions and name regularity experienced during training, making it difficult to generalize unseen entities, especially, entities with rare patterns (e.g., COVID-19).
To support our hypothesis and see if such bias can be handled, we adopt a bias product method \cite{clark-etal-2019-dont}, which is a kind of debiasing method effective in alleviating dataset biases in various NLP tasks such as visual question answering and natural language inference.

\begin{table}[t]
\caption{
Performance of \biobert~and \bert~with/without our debiasing method on \ncbi, \cdrdis, and \cdrchem. 
\textcolor{ao}{$\uparrow$} and \textcolor{cadmiumred}{$\downarrow$} indicate performance increases and decreases when using the method, respectively.
The best scores are highlighted in bold.
}
\label{tab:debiasing}
\begin{adjustbox}{max width = \columnwidth}
\centering
\normalsize
\begin{tabular}{l|lll|lll}
\toprule
 \multicolumn{1}{c}{} & \multicolumn{3}{c}{\textbf{Overall}} & \multicolumn{3}{c}{\textbf{In-depth}} \\
 \cmidrule(lr){2-4} \cmidrule(lr){5-7}
  \multicolumn{1}{l}{\textbf{Model}} & \multicolumn{1}{c}{\textbf{\precision}} & \multicolumn{1}{c}{\textbf{\recall}} & \multicolumn{1}{c}{\textbf{\fscore}} & \multicolumn{1}{c}{\textbf{\mem}} & \multicolumn{1}{c}{\textbf{\syn}} & \multicolumn{1}{c}{\textbf{\con}} \\
\midrule
\multicolumn{7}{c}{\ncbi} \\
\midrule
\biobert  & \textbf{86.7} & \textbf{90.5} & \textbf{88.6} & \textbf{95.5} & 80.9 & 84.1    \\
\quad + Debias.  & 85.0 & 90.2 & 87.5 & 94.2 \textcolor{cadmiumred}{$\downarrow$} & \textbf{81.5} \textcolor{ao}{$\uparrow$} & \textbf{86.1} \textcolor{ao}{$\uparrow$} \\
\midrule
\bert  & 83.8 & 87.6 & 85.6 & 95.0 & 79.0 & 70.7  \\
\quad + Debias.  & 82.0 & 87.0 & 84.5 & 93.1 \textcolor{cadmiumred}{$\downarrow$} & 80.0 \textcolor{ao}{$\uparrow$} & 73.2 \textcolor{ao}{$\uparrow$} \\ 
\midrule
\multicolumn{7}{c}{\cdrdis} \\
\midrule
\biobert & \textbf{82.4} & \textbf{86.3} & \textbf{84.3} & \textbf{93.2} & 74.9 & 73.7  \\
\quad + Debias.  & 80.4 & \textbf{86.3} & 83.3 & 92.3 \textcolor{cadmiumred}{$\downarrow$} & \textbf{77.1} \textcolor{ao}{$\uparrow$} & \textbf{74.7} \textcolor{ao}{$\uparrow$} \\
\midrule
\bert & 78.5 & 81.4 & 79.9 & 91.5 & 64.0 & 63.4 \\
\quad + Debias. & 75.7 & 81.0 & 78.3 & 89.9 \textcolor{cadmiumred}{$\downarrow$} & 66.2 \textcolor{ao}{$\uparrow$} & 64.7 \textcolor{ao}{$\uparrow$} \\
\midrule
\multicolumn{7}{c}{\cdrchem} \\
\midrule
\biobert & \textbf{92.1} & \textbf{93.1} & \textbf{92.6} & \textbf{97.8} & 82.1 & 87.0  \\
\quad + Debias. & 91.2 & \textbf{93.1} & 92.1 & 97.1 \textcolor{cadmiumred}{$\downarrow$} & \textbf{82.8} \textcolor{ao}{$\uparrow$} & \textbf{88.2} \textcolor{ao}{$\uparrow$} \\
\midrule
\bert & 89.8 & 90.0 & 89.9 & 97.0 & 72.4 & 81.1  \\
\quad + Debias. & 87.3 & 90.7 & 89.0 & 96.6 \textcolor{cadmiumred}{$\downarrow$} & 73.7 \textcolor{ao}{$\uparrow$} & 83.9 \textcolor{ao}{$\uparrow$} \\ 
\bottomrule
\end{tabular}
\end{adjustbox}
\end{table}

\begin{table*}[t]
\caption{
Disease entities with rare surface forms and the performance of \biobert~with/without our debiasing method.
}
\label{table:name_reg_results}
\centering
\footnotesize
\begin{tabularx}{0.99\textwidth}{ lc *{9}{Y} }
\toprule
\textbf{Test Entity} & \textbf{COVID-19} & \textbf{47, XXY} & \begin{tabular}[c]{@{}c@{}}\textbf{African Iron} \\ \textbf{Overload} \end{tabular} & \textbf{Bejel} & \begin{tabular}[c]{@{}c@{}}\textbf{Geographic} \\ \textbf{Tongue} \end{tabular} & \textbf{Pinta} &\begin{tabular}[c]{@{}c@{}}\textbf{PMM2} \\ \textbf{-CDG}\end{tabular} & \begin{tabular}[c]{@{}c@{}}\textbf{Precocious} \\ \textbf{Puberty}\end{tabular} & \begin{tabular}[c]{@{}c@{}}\textbf{VACTERL} \\ \textbf{Association}\end{tabular} \\ \midrule
Frequency & 5,287\smallspace & 1,292\smallspace\tinyspace & \smallspace\tinyspace39 & \tinyspace112 & \tinyspace391 & \tinyspace124 & \tinyspace370 & 6,903\smallspace & \tinyspace685 \\ \midrule
\biobert & 45.7 & 1.0 & \smallspace6.2 & 45.5 & 66.8 & 50.6 & 29.0 & 71.1 & 31.0 \\
\quad+ Debias. & 47.3 & 2.1 & 17.4 & 49.8 & 76.5 & 55.0 & 34.6 & 90.0 & 43.8 \\ 
\midrule
Improvement & \textcolor{ao}{\textbf{+ 1.6}}\tinyspace &  \textcolor{ao}{\textbf{+ 1.1}}\smallspace\tinyspace & \textcolor{ao}{\textbf{+ 11.2}}\smallspace & \textcolor{ao}{\textbf{+ 4.3}}\tinyspace & \textcolor{ao}{\textbf{+ 9.7}}\tinyspace & \textcolor{ao}{\textbf{+ 4.4}}\tinyspace & \textcolor{ao}{\textbf{+ 5.6}}\tinyspace & \textcolor{ao}{\textbf{+ 18.9}}\smallspace\tinyspace & \textcolor{ao}{\textbf{+ 12.8}}\smallspace\tinyspace \\ 
\bottomrule
\end{tabularx}
\end{table*}

\subsubsection{Formulation}

Bias product \cite{clark-etal-2019-dont} trains an \emph{original} model using a \emph{biased} model such that the original model does not learn much from spurious cues.
Let $\mathrm{p}_{(n,i)} \in \mathcal{R}^{{K}}$ be the probability distribution over ${K}$ target classes of the original model at the $i$-th word in the $n$-th sentence, and $\mathrm{b}_{(n,i)} \in \mathcal{R}^{{K}}$ be that of the biased model.
We add $\mathrm{log}(\mathrm{p}_{(n,i)})$ and $\mathrm{log}(\mathrm{b}_{(n,i)})$ element-wise, and then calculate a new probability distribution $\hat{\mathrm{p}}_{(n,i)} \in \mathcal{R}^{{K}}$ by applying the softmax function over ${K}$ classes as follows:
\begin{equation}
    \hat{\mathrm{p}}_{(n,i)} = \text{softmax}(\mathrm{log}(\mathrm{p}_{(n,i)}) + \mathrm{log}(\mathrm{b}_{(n,i)})).
\end{equation}

We minimize the negative log-likelihood between the combined probability distribution $\hat{\mathrm{p}}_{(n,i)}$ and the ground-truth label.
This assigns low training signals to words with highly skewed class distributions.
As a result, it prevents the original model from being biased towards statistical cues in datasets.
Note that only the original model is updated, and the biased model is fixed during training.
At inference, we use only the probability distribution of the original model, $\mathrm{p}_{(n,i)}$.

In previous works, biased models are usually pretrained neural networks using hand-crafted features as input \cite{clark-etal-2019-dont, he-etal-2019-unlearn, karimi-mahabadi-etal-2020-end, utama-etal-2020-mind}.
On the other hand, \cite{ko-etal-2020-look} used data statistics as the probability distributions of the biased model, which is computationally efficient and performs well.
Similarly, we calculate the class distribution of each word using training sets, and then use the statistics.
The probability that our biased model predicts $k$-th class is defined as follows:
\begin{equation}
    b^{k}_{(n,i)} = \frac{\sum^{{N}}_{m=1}
    \sum^{{L}_{m}}_{j=1} \mathds{1}_{\vert x_{(m, j)}
    = x_{(n,i)} {\displaystyle \wedge} y^{k}_{(m, j)} =
    1\vert}}{\sum^{{N}}_{m=1}
    \sum^{{L}_{m}}_{j=1} \mathds{1}_{|x_{(m, j)} =
    x_{(n,i)}|
    }},
\end{equation}
where ${N}$ is the number of sentences in the training set, ${L}_m$ is the length of the $m$-th sentence, and $x_{(n,i)}$ is the $i$-th word in the $n$-th sentence.
If the ground-truth label of the word $x_{(n,i)}$ is the $k$-th class, $y^{k}_{(n, i)}$ = 1, otherwise 0.

\subsubsection{Effect of Debiasing}
\label{subsec:effect_of_debiasing}

We explore how the debiasing method affects models' generalization abilities.
\Cref{tab:debiasing} shows models' performance changes after applying the debiasing method.
The method decrease the memorization because it debiases models' bias towards memorizable mentions and their class distributions.
On the other hand, the method constantly improves the performance on \syn~and \con~on the benchmarks.
Debiasing methods usually decrease overall performance on benchmarks \cite{he-etal-2019-unlearn,utama-etal-2020-mind}, which is consistent with our results.
With recent efforts to reduce bias while maintaining overall performance \cite{utama-etal-2020-mind}, our debiasing method could be improved in future work.
Also, the debiasing method changes the model's behavior and corrects the errors in the first and third examples in \Cref{table:evidence_of_bias}.

We also see if our debiasing method can improve the generalizability to entities with weak name regularity.
Before testing the method, we crawled a list of rare diseases and their descriptions from the NORD (National Organization for Rare Disorders) database \footnote{\href{https://rarediseases.org/rare-diseases}{https://rarediseases.org/rare-diseases}} based on our hypothesis that rare diseases tend to have more unique surface forms than common diseases.
Disease names were filtered if \biobert~trained on \ncbi~successfully extracted them based on the descriptions.
Since descriptions provide sufficient context to recognize entities, e.g., ``\textit{African iron overload is a rare disorder characterized abnormally elevated levels of iron in the body},'' an entity's surface form would be rare if a model fails to recognize the entity from the description. 
Thus, we assumed that the remaining diseases after filtering have weak name regularity.
Finally, we obtained 8 diseases from the database and collected PubMed abstracts in which the diseases appear.
\Cref{table:name_reg_results} shows the list of diseases and their frequency of occurrence.
All diseases are different from conventional patterns, and their CUIs are unseen based on the \ncbi~training data.
We tested our debiasing model on the diseases along with COVID-19 using the same relaxed version of recall as the same as for COVID-19.
As a result, our debiasing method consistently improved the generalization to rare patterns as shown in \Cref{table:name_reg_results}.

\subsubsection{Side Effects of Debiasing}
\label{appendix:case_study}

\begin{table}[t]
\caption{Side effects of debiasing.
Entity mentions (ground-truth labels) are displayed in blue. 
Model predictions are highlighted with yellow boxes.
}
\label{tab:side_effects}
\normalsize
\centering
\begin{adjustbox}{max width = \columnwidth}

\begin{tabular}{l}
\toprule
\textbf{Example}  \\ \midrule
\textit{Not noun phrase} ($15/23$)  \\ \midrule
\begin{tabular}[c]{@{}l@{}}  
\relax[1] \colorbox{yellow}{\textbf{Loss of}} righting ability was scored at $\dots$ \\ 
\relax[2] $\dots$ a target to protect \colorbox{yellow}{\textbf{infarcted}} myocardium. \\    
\end{tabular}    \\ \midrule
\textit{w/o Name regularity} $(5/23)$  \\ \midrule
\begin{tabular}[c]{@{}l@{}}
\relax[1] \textbf{{\textcolor{ao}{\colorbox{yellow}{Takotsubo syndrome}}}} (or \textbf{{\textcolor{ao}{\colorbox{yellow}{apical ballooning}}}} \textbf{{\textcolor{ao}{syndrome}}}) \\ $\quad$\smallspace secondary to Zolmitriptan. \\
\relax[2] $\dots$ upregulation of \textbf{{\textcolor{ao}{\colorbox{yellow}{kidney} injury}}} molecule (KIM)-1 $\dots$ \\  
\end{tabular} \\  
\midrule
\textit{Special symbol} $(3/23)$  \\  \midrule
\begin{tabular}[c]{@{}l@{}}
\relax[1] $\dots$ a wide range of {\textcolor{ao}{\colorbox{yellow}{\textbf{cancers}}}} including \textbf{{\textcolor{ao}{\colorbox{yellow}{sarcomas,}}}} \\ $\quad$\smallspace \textbf{{\textcolor{ao}{\colorbox{yellow}{lymphoma}}}}, \textbf{{\textcolor{ao}{\colorbox{yellow}{gynecologic and testicular cancers}}}}. \\ 
\relax[2] Global longitudinal (GLS), circumferential \colorbox{yellow}{\textbf{(}}GCS\colorbox{yellow}{\textbf{)}}, and \\
$\quad$\smallspace radial strain (GRS) were $\dots$ \\
\end{tabular}   \\ 
\bottomrule
\end{tabular}
\end{adjustbox}
\end{table}

Our debiasing method prevents models from overtrusting the class distributions and surface forms of mentions, making the models sometimes predict spans of text as entities, which have never appeared in the training set.
Although this \emph{exploration} of debiased models helps find unseen mentions as shown \Cref{tab:debiasing} and \Cref{table:name_reg_results}, they have some side effects at the expense of the exploration.
To analyze this, we sample 100 cases from the test set of \cdrdis~that an original \biobert~model predicted correctly, but a debiased one failed.

Among all cases, we find 23 abnormal predictions of the debiased model and classify them into three categories as shown in \Cref{tab:side_effects}.
The most frequent type is predicting spans that are not noun phrases.
As shown in the first example in the table, although ``Loss of'' is an incomplete phrase, the model predicted it.
Also, the model predicted the word ``infarcted'' as an entity although the word is an adjective and is only labeled as ``O'' in the training set.
Also, the second type is related to name regularity.
We found that the model sometimes excluded strong patterns from their predictions.
For instance, as shown in \Cref{tab:side_effects}, the model predicted entities without ``syndrome'' and ``injury''.
When using the debiasing method, there can be a trade-off between performance for entities with weak name regularity and those with strong name regularity.
Lastly, the model occasionally predicts special symbols.
As shown in the last row of the table, the model predicted the word ``sarcomas'' with a comma.
The model also recognized parentheses as entities.
From these results, we conclude that the debiasing method can lead to abnormal predictions by encouraging models to predict rare (or never appeared) classes of words and spans during training.

\section{Related Work}

\subsection{BioNER Models}
In recent years, BioNER has received significant attention for its potential applicability to various downstream tasks in biomedical information extraction.
Traditional methods in BioNER are based on hand-crafted rules \cite{fukuda1998toward, proux1998detecting, narayanaswamy2002biological} or biomedical dictionaries \cite{hettne2009dictionary, gerner2010linnaeus}.
However, these methods require the knowledge and labour of domain experts and are also vulnerable to unseen entity mentions. 
With the development of deep learning and the advent of large training data, researchers shifted their attention to neural models \cite{sahu-anand-2016-recurrent, habibi2017deep}, which are based on recurrent neural networks (RNNs) with conditional random fields (CRFs) \cite{lafferty2001conditional}.
These models automatically learn useful features in datasets without the need of human labour and achieve competent performance in BioNER.
The performance of BioNER models has been further improved with the introduction of multi-task learning on multiple biomedical corpora \cite{crichton2017neural, wang2019cross, yoon2019collabonet}.
Several works demonstrated the effectiveness of jointly learning the BioNER task and other biomedical NLP tasks \cite{leaman2016taggerone, watanabe2019multi, zhao2019neural, peng-etal-2020-empirical}.
Recently, pretrained language models such as \biobert~achieved SOTA results in many tasks such as relation extraction and question answering, and also in BioNER \cite{lee2020biobert,peng2019transfer,gu2020domain}.

\subsection{Generalization to Unseen Mentions}
Generalization to unseen mentions has been an important research topic in the field of NER \cite{augenstein2017generalisation, taille2020contextualized,agarwal2020entity}.
Despite recent attempts to analyze the generalization of NER models in the general domain \cite{fu2020rethinking, fu2020interpretable, lin-etal-2020-rigorous, agarwal2021interpretability}, there are few studies in the biomedical domain.
Several studies investigated transferability of BioNER models across datasets \cite{giorgi2018transfer, giorgi2020towards}.
On the other hand, we study the generalization to new and unseen mentions based on our new data partitioning method.
Note that they did not split benchmarks and evaluated models based on overall performance, so our method can be applied to their experimental setups in future work.

\subsection{Dataset Bias}
While many recent studies pointed out dataset bias problems in various NLP tasks such as sentence classification \cite{poliak-etal-2018-hypothesis, niven-kao-2019-probing, mccoy-etal-2019-right} and visual question answering \cite{agrawal2018don}, neither works raise bias problems regarding BioNER benchmarks.
Our work is the first to deal with dataset bias in BioNER and to demonstrate the effectiveness of the debiasing method.
Recent works found that \emph{low} label consistency (the degree of label agreement of an entity on the training set) decreases the performance of models on general NER benchmarks \cite{fu2020rethinking, fu2020interpretable}.
In this work, we show that \emph{high} label consistency also can harm the generalization when the label distribution of the test set is different from that of the training set.

\section{Conclusion}
\label{sec:conclusion}

In this work, we thoroughly explored the memorization, synonym generalization, and concept generalization abilities of existing BioNER models.
We found current best NER models are overestimated, tend to rely on dataset biases, and have difficulty recognizing entities with novel surface patterns. 
Finally, we showed that the generalizability can be improved using a current debiasing method.
We hope that our work can provide insight into the generalization abilities of BioNER models and new directions for future work.

\appendix

\section{Details in Partitioning Benchmarks}
\label{appendix:partitioning}

We classify the set of mentions $\{e_{(n, t)}$: $e_{(n, t)}$ $\in$ $\mathbb{E}_\text{train}$, $c_{(n, t)}$ $\notin$ $\mathbb{C}_{\text{train}}\}$ into \mem~for single-type datasets (e.g., \ncbi~and {CDR}).
If a dataset is multi-type (e.g., {MedMentions}), we classify the mentions into \con.
Since there are entity mentions that are mapped to more than one CUI, $c_{(n,t)}$ does not have to be a single CUI and may be a list of CUIs.
In this case, we classify the mentions into \con~if all CUIs in the list are not included in $\mathbb{C}_\text{train}$ and otherwise as \syn.
We classify mentions with the unknown CUI ``-1'' into \con~because unknown concepts in the training and test sets are usually different.
We lowercase mentions and remove punctuation in them when partitioning benchmarks.

\section{Model Comparison}
\label{appendix:model_description}

Our neural baseline models (i.e., BERT, BioBERT, BlueBERT, and PubMedBERT) have the same model architecture, which are Transformer-based encoders~\cite{vaswani2017attention} with a linear classifier.
They differ in vocabulary, initialization method, and training corpus during pre-training, as summarized in~\Cref{tab:difference_models}.
First, BERT is trained on Wikipedia and the BookCorpus~\cite{zhu2015aligning} from scratch using the vocabulary within the corpora.
BioBERT and BlueBERT are initialized with BERT's weights and further trained on PubMed articles.
Additionally, BlueBERT is trained on the MIMIC-III corpus, which consists of clinical notes.
PubMedBERT is also trained on the PubMed corpus, but it is trained from scratch and trained with the PubMed vocabulary.

\section{Implementation Details}
\label{appendix:implementation}

\begin{table}[t]
\caption{Differences between pretrained language models. 
Vocab. and Init. indicate vocabulary and initialization.
}
\label{tab:difference_models}
\centering
\normalsize
\begin{adjustbox}{max width = \columnwidth}
\begin{tabular}{llll}
\toprule
\textbf{Model} & \textbf{Vocab.} & \textbf{Init.} & \textbf{Corpus} \\
\midrule
\pubmedbert~\cite{gu2020domain} & PubMed & - & PubMed \\
\bluebert~\cite{peng2019transfer} & Wiki+Books  & BERT & PubMed+MIMIC \\
\biobert~\cite{lee2020biobert}  & Wiki+Books & BERT & PubMed \\
\bert~\cite{devlin-etal-2019-bert} & Wiki+Books & - & Wiki+Books  \\
\bottomrule
\end{tabular}
\end{adjustbox}
\end{table}

\begin{table}[t]
\caption{
Best configurations of model hyperparameters.
}
\label{tab:hyperparams}
\begin{adjustbox}{max width = \columnwidth}
\centering
\footnotesize
\begin{tabular}{llcc}
\toprule
  \multicolumn{1}{l}{\textbf{Dataset}}  & \multicolumn{1}{l}{\textbf{Model}}  & \multicolumn{1}{l}{\begin{tabular}{c} \textbf{Learning} \\ \textbf{Rate} \end{tabular}} & \textbf{Temperature}  \\
 \midrule
 \multirow{6}{*}{\ncbi} & \pubmedbert & 5e-5 & - \\
& \bluebert & 5e-5 & - \\
& \biobert & 5e-5 & \textit{none} \\
 & \biobert + Debias. & 3e-5 & \textit{none} \\
 & \bert & 5e-5 & - \\
 & \bert + Debias. & 3e-5 & 1.1 \\
 \midrule
 \multirow{6}{*}{\cdrdis} & \pubmedbert & 5e-5 & - \\
 & \bluebert & 5e-5 & - \\
 & \biobert  & 5e-5 & - \\
 & \biobert + Debias. & 5e-5 & 1.1 \\
 & \bert & 5e-5 & - \\
 & \bert + Debias & 5e-5 & 1.1 \\
 \midrule
 \multirow{6}{*}{\cdrchem} & \pubmedbert & 5e-5 & - \\
& \bluebert & 5e-5 & - \\
& \biobert & 5e-5 & -  \\
 & \biobert + Debias. & 5e-5 & 1.1  \\
 & \bert & 3e-5 & - \\
 & \bert + Debias. & 1e-5 & 1.1  \\
\bottomrule
\end{tabular}
\end{adjustbox}
\end{table}

In the experiments, we used a public PyTorch implementation provided by \cite{lee2020biobert}.\footnote{\href{https://github.com/dmis-lab/biobert-pytorch}{https://github.com/dmis-lab/biobert-pytorch}}
We used the bert-base-cased model for BERT,\footnote{\href{https://huggingface.co/bert-base-cased}{https://huggingface.co/bert-base-cased}} the biobert-base-cased-v1.1 model for \biobert,\footnote{\href{https://github.com/dmis-lab/biobert}{https://github.com/dmis-lab/biobert}} the bluebert\_pubmed\_uncased\_L-24\_H-1024\_A-16 model for BlueBERT,\footnote{\href{https://huggingface.co/bionlp/bluebert\_pubmed\_uncased\_L-24\_H-1024\_A-16}{https://huggingface.co/bionlp/bluebert\_pubmed\_uncased\_L-24\_H-1024\_A-16}} and the BiomedNLP-PubMedBERT-base-uncased-abstract model for PubMedBERT.\footnote{\href{https://huggingface.co/microsoft/BiomedNLP-PubMedBERT-base-uncased-abstract}{https://huggingface.co/microsoft/BiomedNLP-PubMedBERT-base-uncased-abstract}}
The max length of input sequence is set to 128.
Sentences whose lengths are over 128 are divided into multiple sentences at the preprocessing stage.
We trained and tested our models on a single Quadro RTX 8000 GPU.
For our synonym dictionaries, we used the July 2012 version of MEDIC~\cite{davis2012medic} and the November 2019 version of CTD (Comparative Toxicogenomics Database), provided by~Sung~\etal~\cite{sung-etal-2020-biomedical}.

For all models, we used the batch size of 64 and searched learning rate in the range of \{1e-5, 3e-5, 5e-5\}.
For our debiasing method, we smooth the probability distribution of the biased model using temperature scaling \cite{guo2017calibration} since excessive penalties for bias can hinder the learning process.
We searched the scaled parameter in the range of \{\textit{none}, 1.1\}, where \textit{none} indicates that temperature scaling is not applied.
We chose the best hyperparameters based on the F1 score on the development set.
The selected hyperparameters are described in \Cref{tab:hyperparams}.
Note that all results are averaged over 5 runs using a randomly selected seed.

The original \biobert~model \cite{lee2020biobert} was trained on not only the training set, but also the development set, after the best hyperparameters are chosen based on the development set.
This approach improves performance in general when the number of training examples is insufficient and is commonly used in many studies in BioNER.
On the other hand, we did not use the development set for training models, resulting in lower performance of \bert~and \biobert~compared to the performance reported by Lee~\etal~\cite{lee2020biobert}.

\section{Annotation Inconsistency in Biomedical Databases}
\label{appendix:annotation_inconsistency}
As shown in \Cref{tab:main_result}, the performance on \mem~of \dictsyn~is lower than that of \dicttrain~as there exists annotation inconsistency between benchmarks and biomedical databases.
For example, ``seizures'' and ``generalized seizures'' are entities with the same concept in the databases, so the dictionary of \dictsyn~includes both ``seizures'' and ``generalized seizures.''
However, in \cdrdis~only ``seizures'' is annotated. 
Since dictionary models predict the longest text spans that are in their dictionaries, \dictsyn~predicts ``generalized seizures,'' resulting in incorrect prediction.
Also, the dictionary models cannot generalize to new concepts, but \dictsyn~achieved recall of 1.4 on \con~of \cdrdis~due to annotation inconsistency, i.e., there are mentions with the same surface forms, but different CUIs.

\section{Tokenization Issue}
Following Lee~\etal~\cite{lee2020biobert}, we split words into subwords based on punctuations.
For example, ``COVID-19'' is splitted into three words ``COVID'', ``-,'' and ``19''.
This tokenization makes it easy to deal with nested entities.
If ``SARS-CoV'' is splitted into subwords,  a model cannot detect ``SARS'' as a disease.
However, the tokenization is not an optimal way in predicting the whole word ``SARS-CoV'' as a virus.

To see if dramatic low performance is due to the tokenization issue, we preprocessed ``COVID-19'' as a single word and tested \biobert~on them.
As a result, the performance of \biobert~has improved from 45.7 to 54.1, and from 3.4 to 15.4, when trained on \ncbi~and \cdrdis, respectively.
Although the change in tokenization clearly boosts performance, we have seen that the performance improvement in \Cref{tab:edit_covid_form} and \Cref{tab:augment_results}, which is not explained by the tokenization issue alone.
The main reason for the failure in recognizing COVID-19 is that models are vulnerable to unique name patterns.

\section*{Acknowledgments}
We would like to thank Jinhyuk~Lee, Mujeen~Sung, Minbyul~Jeong, Sean~Yi, Gangwoo~Kim, Wonjin~Yoon, and Donghee~Choi for the helpful feedback.

\clearpage

\bibliography{bibliography}
\bibliographystyle{unsrt}

\EOD

\end{document}